\documentclass{article}

\usepackage[preprint]{nips_2018}

\usepackage[utf8]{inputenc} 
\usepackage[T1]{fontenc}    
\usepackage{hyperref}       
\usepackage{url}            
\usepackage{booktabs}       
\usepackage{amsfonts}       
\usepackage{nicefrac}       
\usepackage{microtype}      

\usepackage{varwidth}
\usepackage{capt-of}
\usepackage{lipsum}
\usepackage{wrapfig}
\usepackage{sidecap}
\usepackage{pifont}
\usepackage{comment}

\usepackage{multirow}

\usepackage{amsmath}
\DeclareMathOperator*{\argmin}{argmin}
\DeclareMathOperator*{\argmax}{argmax}

\usepackage{graphicx}

\title{Mapping Unparalleled Clinical Professional and Consumer Languages with Embedding Alignment}

\author{
  Wei-Hung Weng {\normalfont and} Peter Szolovits \\
  Computer Science and Artificial Intelligence Laboratory \\
  Massachusetts Institute of Technology \\
  Cambridge, MA, USA 02139 \\
  \texttt{\{ckbjimmy,psz\}@mit.edu} \\
}

\begin{document}

\maketitle

\begin{abstract}
Mapping and translating professional but arcane clinical jargons to consumer language is essential to improve the patient-clinician communication.
Researchers have used the existing biomedical ontologies and consumer health vocabulary dictionary to translate between the languages.
However, such approaches are limited by expert efforts to manually build the dictionary, which is hard to be generalized and scalable.
In this work, we utilized the embeddings alignment method for the word mapping between unparalleled clinical professional and consumer language embeddings.
To map semantically similar words in two different word embeddings, we first independently trained word embeddings on both the corpus with abundant clinical professional terms and the other with mainly healthcare consumer terms.
Then, we aligned the embeddings by the Procrustes algorithm. 
We also investigated the approach with the adversarial training with refinement.
We evaluated the quality of the alignment through the similar words retrieval both by computing the model precision and as well as judging qualitatively by human.
We show that the Procrustes algorithm can be performant for the professional consumer language embeddings alignment, whereas adversarial training with refinement may find some relations between two languages.
\end{abstract}

\section{Introduction}
Better patient-clinician communication is necessary to prevent from defensive medicine or overtreatment~\cite{omer2013impact}.
In the common clinical setting, clinicians heavily use jargons and abbreviations to record patients' clinical history, condition, progression and results of examinations---this is convenient and time-saving for fast communication between clinicians.
While discharging patients, clinicians usually provide a clinical summary along with the discharge instruction in consumer-level language for patients and their family.
However, the instruction includes very limited information that might not be able to represent the patient's clinical status and disease progression.
The consumers may not obtain full information through these materials.
To understand more about their clinical conditions for decision makings, for example, seeking for the second opinion of treatment, it is inevitable to dive into the professional-level sections of clinical summary.
Yet without domain knowledge and training, consumers may have a hard time to clearly understand details through the professional-level clinical summary, especially the domain-specific information recorded by clinical specialists.
For example, parsing the sentence {\tt ``On floor pt found to be hypoxic on O2 4LNC O2 sats 85 \%, CXR c/w pulm edema, she was given 40mg IV x 2, nebs, and put out 1.5 L UOP, she was also put on a NRB with improvement in O2 Sats to 95 \%''} is not difficult to understand for a trained clinician yet nearly the Voynich manuscript for consumers.
Therefore, how to translate clinical professional language to consumer-level language is essential to improve the communication between consumers and clinicians, as well as to assist consumers' decision makings.

Recent studies demonstrate that explaining the same clinical condition with either clinical professional or consumer-level language affects consumers' decision makings~\cite{nickel2017words,omer2013impact}. 
Breast cancer patients tend to accept aggressive surgical treatment if they are told that their lesion is ``cancer'' rather than ``lesion'' or ``abnormal cells'' are found~\cite{omer2013impact}. 
For example, gynecological patients who received the diagnosis of ``polycystic ovary syndrome'' (PCOS) tend to accept more examination, such as the ultrasonography, when the professional term ``PCOS'' rather than ``hormone imbalance'' was informed---although PCOS is indeed a kind of female hormone imbalance~\cite{copp2017influence}.
Such unnecessary overdiagnosis/overtreatment decisions may come from the anxiety and fear about unknown medical domain knowledge for the professional language.
The huge information gap may further yield a potential conflict between patients and clinicians and eventually result in defensive medicine and overtreatment. 

For effective and suitable clinical decision makings, researchers attempted to automatically map and translate clinical professional terms to appropriate consumer terms in clinical narrative texts using existing biomedical ontologies and dictionaries~\cite{zielstorff2003controlled}.
Zeng-Treitler et al. mapped clinical texts to be comprehensible to non-professionals using the Unified Medical Language System (UMLS) Metathesaurus with the consumer health vocabulary (CHV) to perform the synonym replacement and medical concept explanation insertion for mapping and translation~\cite{zeng2007making}. 
Alfano et al. also adopted CHV to build the prototype of medical term mapping~\cite{alfano2015automatic}.
However, the evidence shows that the ontology and dictionary mapping-based approaches have a limitation---it may not be able to map the terms outside the vocabulary space of CHV~\cite{keselman2008consumer}.
Vydiswaran et al. used the pattern-based mining on Wikipedia database to explore the mapping between professional and consumer languagues~\cite{vydiswaran2014mining}.
The pattern-based approach is more generalized, yet the information from Wikipedia may not be the appropriate proxy of professional language that physicians commonly use in the real clinical setting.

The advances of machine learning and natural language processing (NLP) have been utilized to perform different clinical NLP tasks, from word, sentence to document-level classification~\cite{dernoncourt2017identification,lee2016sequential,weng2017medical}.
For word-level NLP tasks, unsupervised word embeddings techniques learn continuous-valued word vector representations through the co-occurrence information~\cite{mikolov2013distributed,pennington2014glove,bojanowski2017enriching}.
These techniques have become one of the standard approaches to find the word similarity.
The concept of learning word embeddings has also been extended to learning embeddings alignment for unimodal cross-lingual, and even cross-modal translation with minimal human supervision~\cite{conneau2018word,chung2018cross}.
Utilizing machine learning and NLP techniques for clinical NLP problems to align professional and consumer language embedding spaces may have a potential to map and translate the languages with minimal supervision and scalability.

In this work, we investigated the feasibility of learning the embeddings alignment between two characteristically different clinical corpora.
We trained two word embeddings independently on the clinical note sections with abundant clinically professional jargons and the others with consumer colloquial terms~\cite{mikolov2013distributed,bojanowski2017enriching}.
Next, we investigated both the Procrustes algorithm with anchors, and the adversarial training with refinement for the unparalleled mapping between two embedding spaces~\cite{conneau2018word,lample2018unsupervised}.
Finally, we evaluated the quality of the aligned embeddings through the similar word retrieval task. 
To our knowledge, this is the first work that applies the embedding alignment approach for the unparalleled mapping between clinical professional and consumer languages.

\section{Methods}

\subsection{Learning Word Embeddings}
We adopted the word-level and subword-level skip-grams algorithms for learning word embeddings of both clinical professional and consumer languages~\cite{mikolov2013distributed,bojanowski2017enriching}.
In detail, we trained the embeddings by setting the window size~$k=3, 5$.
We considered the words only appear more than 3 times, and the negative sampling rate of $10^{-5}$.
The model was trained by stochastic gradient descent~(SGD) without momentum with a fixed learning rate of~$0.05$ for 20 epochs.
We experimented on the embedding dimension of 200.

\subsection{Embeddings Alignment}
Assuming that we have the $x$-word, $d$-dimension professional language embedding~$\mathcal{P} = \{p_{1}, p_{2}, \ldots, p_{x}\}\subseteq \mathbb{R}^{d}$ and the $y$-word, $d$-dimension consumer language embedding~$\mathcal{C} = \{c_{1}, c_{2}, \ldots, c_{y}\}\subseteq \mathbb{R}^{d}$.
We constructed the synthetic mapping dictionary to learn a linear mapping matrix~$W$ between the two embedding spaces, such that~$p_{i}\in \mathcal{P}$ corresponds to which~$c_{j}\in \mathcal{C}$.
Then we have the following equation:
\begin{align*}
  W^{\star} = \argmin_{W\in \mathbb{R}^{d \times d}} \|WX - Y\|^{2}
\end{align*}
where~$X$ and~$Y$ are two aligned matrices of size~$d \times k$ formed by~$k$-word embeddings selected from~$\mathcal{P}$ and~$\mathcal{C}$.

We further added the orthogonality constraint on~$W$, where the above equation will turn into the Procrustes problem that can be solved by singular value decomposition (SVD) with a closed form solution~\cite{xing2015normalized}:
\begin{equation*}
  W^{\star} = \argmin_{W\in \mathbb{R}^{d \times d}} \|WX - Y\|^{2} = UV^T, \text{ where } U \Sigma V^T = \text{SVD}(YX^T)
\end{equation*}
The aligned output of the professional language input~$a$ will be $\argmax_{c_{j}\in \mathcal{C}}\cos(Wp_{a}, c_{j})$.

To reach the minimal supervision, we did not use any clinical term mapping dictionaries, such as UMLS CHV, in the experiments.
Instead, we leveraged the characteristics of two embeddings, which are both in English, and only used identical character strings in the embeddings to form a synthetic dictionary for learning the mapping matrix $W$.
The identical strings serve as anchors in order to learn $W$ with the iterative Procrustes algorithm.

To search the nearest neighbors (the most similar words), Cross-Domain Similarity Local Scaling (CSLS) was calculated to reduce the effect of the hubness problem that a data point tends to be nearest neighbors of many points in a high-dimensional space~\cite{conneau2018word,dinu2015improving}.

\paragraph{Adversarial Training}
We also experimented with adversarial training in case that no identical strings between embeddings can be found.
We first learn an approximated proxy for $W$ using the generative adversarial network (GAN) to make the aligned~$\mathcal{P}$ and~$\mathcal{C}$ indistinguishable, then refine by the iterative Procrustes algorithm to build the synthetic parallel dictionary~\cite{conneau2018word,goodfellow2014generative} .

In adversarial training, the discriminator aims to discriminate between elements randomly sampled from~$W\mathcal{P} = \{Wp_{1}, Wp_{2}, \ldots, Wp_{x}\}$ and~$\mathcal{C}$.
The generator, $W$, is trained to prevent the discriminator from making an accurate prediction.
Given $W$, the discriminator parameterized by~$\theta_{D}$ try to minimize the following objective function ($\text{Pro}=1$ indicates that it is professional language but not consumer language):
\begin{equation*}
  \mathcal{L}_{D}(\theta_{D}|W) = -\frac{1}{x}\sum_{i = 1}^{x}\log \mathbb{P}_{\theta_{D}}(\text{Pro} = 1 | Wp_{i}) - \frac{1}{y}\sum_{j = 1}^{y}\log \mathbb{P}_{\theta_{D}}(\text{Pro} = 0 | c_{j}).
\end{equation*}
Instead,~$W$ minimizes the following objective function to fool the discriminator:
\begin{equation*}
  \mathcal{L}_{W}(W|\theta_{D}) = -\frac{1}{x}\sum_{i = 1}^{x}\log \mathbb{P}_{\theta_{D}}(\text{Pro} = 0 | Wp_{i}) - \frac{1}{y}\sum_{j = 1}^{y}\log \mathbb{P}_{\theta_{D}}(\text{Pro} = 1 | c_{j})
\end{equation*}
The optimizations are executed iteratively to minimize~$\mathcal{L}_{D}$ and~$\mathcal{L}_{W}$ until convergence~\cite{goodfellow2014generative}.

For the discriminator, we used a two-layer neural network of size 2048 with 10\% neuron dropout, and Leaky ReLU as the activation function.
We trained both the discriminator and~$W$ by SGD with a fixed learning rate of~$10^{-3}$.

The refinement of the matrix $W$ after adversarial training was done by iterative Procrustes algorithm, and CSLS was used to decide mutual nearest neighbors.
We ran 20 iterations of refinement procedure for all experiments.

\section{Experiments}

\subsection{Materials}
\paragraph{Dataset}
Data was collected from the MIMIC-III database~\cite{johnson2016mimic}, which contains 58,976 ICU patients admitted to the Beth Israel Deaconess Medical Center (BIDMC), a large, tertiary medical center in Boston, Massachusetts, USA. 
The database contains detailed information on patients admitted between 2001 and 2012, including hospital administrative data, vital signs, medications, laboratory test results and survival data after hospital discharge.  

We extracted 59,654 free-text discharge summaries from the MIMIC-III database.
For all discharge summaries, we extracted and preprocessed the sections of ``History of present illness'', ``Brief hospital course'', ``Discharge instruction'' and ``Followup instruction''.
We selected the sections of ``History of present illness'' and ``Brief hospital course'' to represent the content with professional jargons since these sections are usually the most narrative components with thoughts and reasoning for the communication between clinicians.
``Discharge instruction'' and ``Followup instruction'' sections instead have consumer-level language and are written for patients and their family. 
For training word embeddings, there are 443,585 sentences in the clinical professional language set and 73,349 sentences in the consumer language set.
There are 26,333 and 6,752 unique words in the professional and consumer term embeddings, respectively.

Although the professional and consumer language set are both from MIMIC, the content in the source and target corpora are not parallel.
However, we expect that there are some overlapping terms in two corpora since both of them are written in English.
We utilized these overlapping English terms as anchors during alignment.

\paragraph{Language Preprocessing}
We applied Stanford CoreNLP toolkit, Natural Language Toolkit (NLTK), and Porter stemming algorithm for common linguistic preprocessing steps, such as clinical document section and sentence fragmentation, word tokenization, stopwords removal and word stemming, before further tasks~\cite{manning2014stanford,bird2004nltk,porter1980algorithm}.

\subsection{Results and Discussion}
We performed the mapping word retrieval task to evaluate the quality of the alignment.
For the ground truth, we used a list of 100 professional-consumer term pairs created by the clinician.
To compute the precision of mapping word retrieval, we queried the nearest $k$ words ($k=1, 5, 10$) from the consumer language embedding using each professional term in the aligned professional language embedding.

To compare the difference between using the smaller clinical corpus and larger general biomedical literature corpus, we used the embedding trained on Pubmed Central Open Access subset (PMC) and PubMed 5.4B-token/2.2M-vocabulary corpora as the baseline of professional language embedding, as well as the embedding trained on 6B-token/400K-vocabulary Wikipedia corpus as the baseline of consumer language embedding~\cite{chiu2016train,pennington2014glove}.


The results of the mapping word retrieval using Procrustes algorithm with anchors approach are shown in Table~\ref{tab:res}.
\begin{table}[ht]
 \centering
\begin{tabular}{c|c|c|c|c|c|c}
  Source & Target & Embedding & Window & P@1 & P@5 & P@10 \\ \hline
  MIMIC-P & MIMIC-C & word & 3/3 & 0.17 & 0.39 & 0.48 \\
  MIMIC-P & MIMIC-C & word & 5/5 & 0.19 & 0.42 & 0.54 \\
  MIMIC-P & MIMIC-C & subword & 3/3 & 0.27 & \textbf{0.57} & \textbf{0.78} \\
  MIMIC-P & MIMIC-C & subword & 5/5 & \textbf{0.30} & 0.55 & 0.68 \\
  PMC-Pubmed & MIMIC-C & word & 30/3 & 0.26 & 0.40 & 0.44 \\
  PMC-Pubmed & MIMIC-C & word & 30/5 & 0.18 & 0.39 & 0.44 \\
  PMC-Pubmed & MIMIC-C & subword & 30/3 & 0.23 & 0.34 & 0.44 \\
  PMC-Pubmed & MIMIC-C & subword & 30/5 & 0.23 & 0.41 & 0.49 \\
  PMC-Pubmed & Wikipedia & word & 30/10 & 0.14 & 0.32 & 0.41 
\end{tabular}
\caption{Performance of mapping word retrieval using Procrustes algorithm. The word-level embedding is derived from the original word2vec skip-grams algorithm, the subword-level embedding is generated by fastText skip-grams algorithm. P@$k$ means the precision at $k$ and P@1 is equivalent to accuracy. MIMIC-P and MIMIC-C represents the professional and consumer language set, respectively.}
\label{tab:res}
\end{table}

Subword-level fastText word embeddings outperform original word-level word2vec embeddings in most cases. 
This is highly likely because that subword-level fastText models utilizes the character-level n-grams information.
Subword-level word embedding is useful in capturing morphological patterns and therefore may enhance the information about word semantics, especially for our mapping word retrieval task.

Even though the MIMIC dataset is much smaller than PMC-Pubmed and Wikipedia, the performance of using MIMIC is better than using larger corpora.
We hypothesized that the discharge summaries from the MIMIC dataset are much suitable to represent the clinical professional and consumer language, comparing with the general PMC-Pubmed and Wikipedia corpora.

\begin{figure}[ht]
  \centering
  \includegraphics[width=0.46\textwidth]{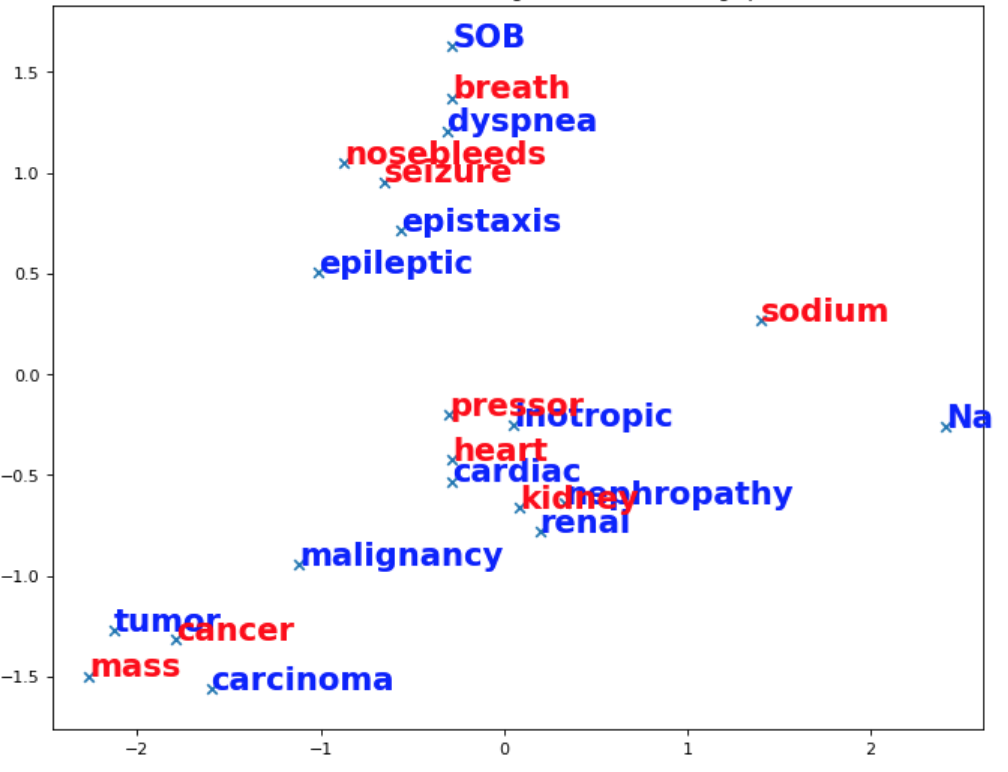}
  \includegraphics[width=0.52\textwidth]{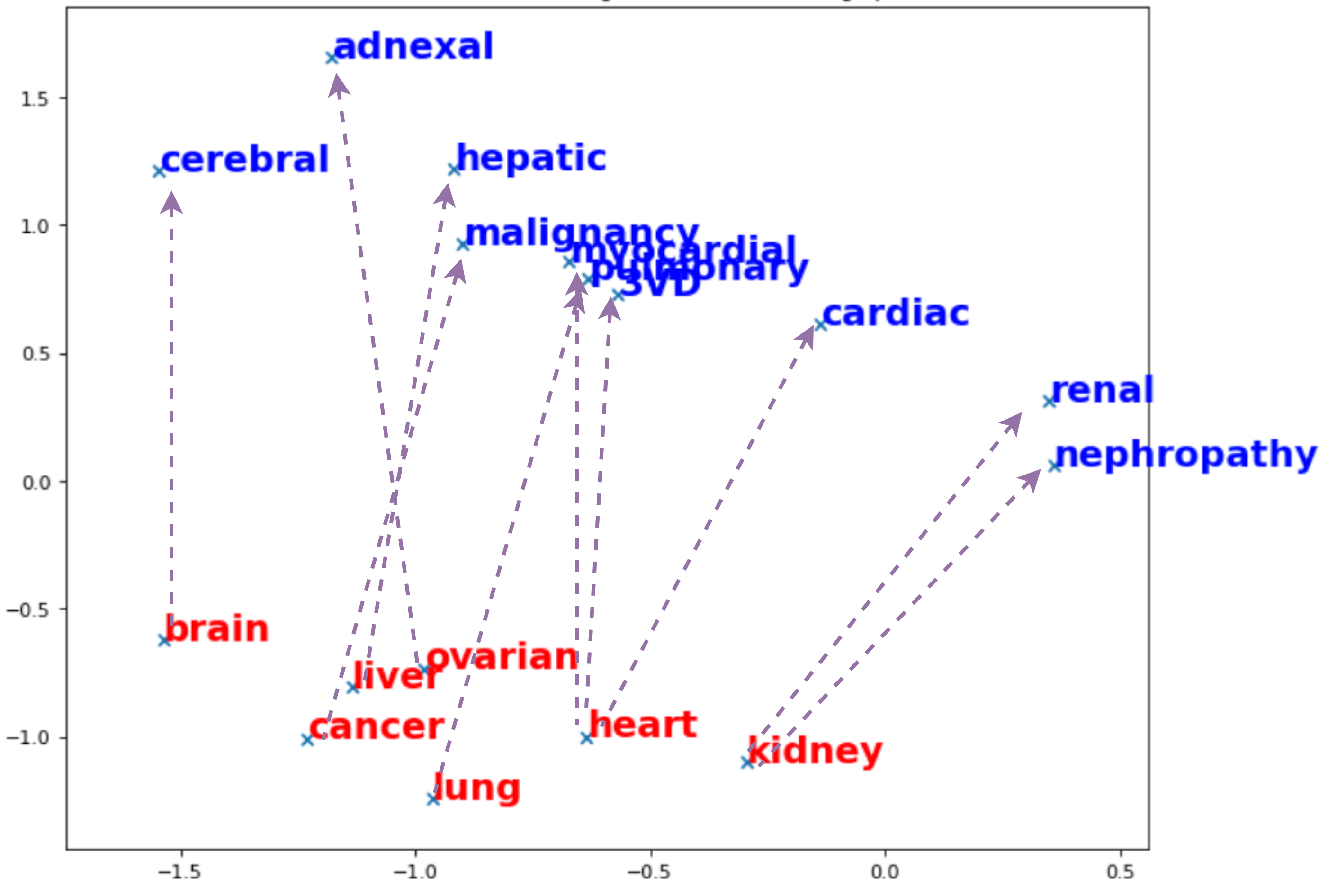}
  \caption{Visualization of the aligned embedding spaces using principal component analysis (PCA). (Left) Procrustes algorithm with anchors approach. The semantically similar professional (blue) and consumer terms (red) are clustered. (Right) Adversarial training approach. There are similar relation vectors between professional (blue) and corresponding consumer anatomy-related terms (red).}
  \label{fig:viz}
\end{figure}

In Table~\ref{tab:similar}, we demonstrate that we retrieve the meaningful mapped consumer terms from the aligned embedding using professional terms as queries through mapping word retrieval.
\begin{table*}[h]
 \centering
 \resizebox{\columnwidth}{!}{
 \begin{tabular}{c|c|c|c|c|c|c|c}
  epistaxis & cardiac & nephropathy & cholangiography & qd & tumor & hepatic & hematemesis \\ \hline
  spontaneous & catheterization & \textbf{renal} & drug-eluting & EC & tumor & \textbf{liver} & \textbf{coffee-ground} \\
  coffee-ground & \textbf{heart} & \textbf{kidney} & \textbf{stent} & \textbf{once/day} & \textbf{cancer} & hepatic & black \\
  light-headed & attack & hepatitis & ureteral & QD & obstructing & Whipple & \textbf{bloody} \\
  \textbf{nosebleed} & coronary & HIV & \textbf{stented} & Ramipril & resection & gastropathy & tarry \\
  stools & cardiac & aka & circumflex & Zocor & \textbf{mass} & Pork & colored \\
  melena & angioplasty & diastolic & bare & meq & metastatic & Mayonnaise & grounds \\
  \textbf{bleeds} & myocardial & pancreatitis & \textbf{stents} & Mesalamine & cerebellum & belly & stools \\
  bloody & cardioversion & Epo & sphincterotomy & 3.125 & occipital & portal & bright \\
  black/tarry & bypass & Diabetes & metal & QHS & hepatocellular & scarring & black/tarry \\
  10days & EP & lupus & \textbf{biliary} & 162 & polypectomy & Y & dark \\
 \end{tabular}
 }
 \caption{Examples of the top-10 nearest neighbor terms in the consumer language embedding queried by clinical professional terms (the topmost row). We highlight the commonly used appropriate corresponding consumer terms of each queried clinical professional term in boldface. For example, ``nosebleed'' is the appropriate consumer version of ``epistaxis''.}
 \label{tab:similar}
\end{table*}
In Figure~\ref{fig:viz}, we visualize the aligned embeddings spaces with principal component analysis (PCA). 
We found that using the Procrustes algorithm with anchors can align two embeddings well. 
The semantically similar clinical professional terms (blue) and consumer (red) terms are close to each other. 
For example, ``mass'', ``tumor'', ``cancer'' ``malignancy'', ``carcinoma'' are clustered~\ref{fig:viz}, and the profession abbreviation ``SOB'' as well as the professional term ``dyspnea'' is close to the consumer term ``breath''.

We did not observe the comparable performance of embeddings alignment in adversarial training approaches.
Possible reasons are due to the training sample size and the possible unsimilar shapes of distribution between the source and target embeddings.
Yet, we observed a pattern that the aligned embeddings generated by the adversarial training capture the relation between professional and consumer anatomy-related terms as shown in Figure~\ref{fig:viz} (right).
Further investigation of the relation vector between the professional-consumer term pairs is required.

\section{Conclusion and Future Work}
We demonstrate the capability of embeddings alignment for mapping unparalleled clinical professional and consumer languages in word-level.
We found that the Procrustes algorithm with anchors approach with the subword-level word embeddings trained on clinical narrative texts, rather than larger general corpus, outperformed the other combinations.
The aligned embeddings learned from the adversarial training approach reveal the relation between professional and consumer anatomy-related terms.

In this study, we first performed clinical professional and consumer language embeddings alignment without the knowledge and supervision from biomedical ontologies and dictionaries, and just use the minimal supervision using the identical strings across corpora.
We also applied the method to the real clinical text corpora, which are derived from the clinical discharge summaries in the MIMIC-III database.

Some limitations in the current study provide the possibility of future directions. 
We need to explore larger clinical note sets to improve the quality of word embeddings and validate the method.
We can also extend the word-level approach to concept-level approach.
The issue of the instability of adversarial training also needs to be considered in the future.
For instance, using Wasserstein GAN~\cite{arjovsky2017wasserstein}, 
or cycle-consistent adversarial networks,~\cite{zhu2017unpaired}, 
instead of original GAN, may be potential approaches to improve the performance.

\bibliographystyle{IEEEtran}
\bibliography{mybib}

\end{document}